\documentclass[conference]{IEEEtran}
\IEEEoverridecommandlockouts
\usepackage{cite}
\usepackage{amsmath,amssymb,amsfonts}
\usepackage{algorithmic}
\usepackage{graphicx}
\usepackage[ruled,vlined]{algorithm2e}
\usepackage{titlesec}

 \makeatletter
\DeclareRobustCommand{\textsupsub}[2]{{%
  \m@th\ensuremath{%
    ^{\mbox{\scriptsize #1}}%
    _{\mbox{\scriptsize #2}}%
  }%
}}
\makeatother

\usepackage{textcomp}
\usepackage{xcolor}
\usepackage{url}
\usepackage{hyperref}
\def\BibTeX{{\rm B\kern-.05em{\sc i\kern-.025em b}\kern-.08em
    T\kern-.1667em\lower.7ex\hbox{E}\kern-.125emX}}

\newcommand{\ts}{\textsuperscript}

\begin{document}

\title{Sliced $\mathcal{L}_2$ Distance for Colour Grading \\

\thanks{This work is partly supported by a scholarship from Umm Al-Qura University, Saudi Arabia, and the ADAPT Centre for Digital Content Technology that is funded under the SFI Research Centres Programme (Grant 13/RC/2106) and is co-funded under the European Regional Development Fund.}
}

\author{\IEEEauthorblockN{Hana Alghamdi}
\IEEEauthorblockA{\textit{Computer Science \& Statistics} \\
\textit{Trinity College Dublin}\\
Dublin, Ireland \\
alghamdh@tcd.ie}
\and
\IEEEauthorblockN{Rozenn Dahyot}
\IEEEauthorblockA{\textit{Computer Science \& Statistics} \\
\textit{Trinity College Dublin}\\
Dublin, Ireland \\
 \href{https://orcid.org/0000-0003-0983-3052}{ORCID:0000-0003-0983-3052}}
 }

\maketitle

\begin{abstract}
We propose a new method with $\mathcal{L}_2$ distance that maps one $N$-dimensional distribution to another, taking into account available information about correspondences. We  solve the high-dimensional problem in 1D space using an iterative projection approach. To show the potentials of this mapping, we apply it to colour transfer between two images that exhibit overlapped scenes. Experiments show quantitative and qualitative competitive results as compared with the  state of the art colour transfer methods.
\end{abstract}

\begin{IEEEkeywords}
$\mathcal{L}_2$, Colour transfer, Colour correction 
\end{IEEEkeywords}

\section{Introduction}

Optimal Transport (OT) has been successfully used as a way for defining cost functions for optimization when performing colour distribution transfer \cite{peyre2019computational}, and have been used more recently to solve machine learning problems  \cite{TanakaNIPS2019,NIPS2019_projPursuit, NIPS2019_Subspace_Detours}. 
Popular OT algorithms such as Iterative Distribution Transfer (IDT) \cite{Pitie_CVIU2007,Pitie_ICCV2005} and Sliced Wasserstein Distance (SWD) \cite{BonneelJMIV2015} use iteratively 
the  OT explicit solution available for mapping 1D projected source dataset onto a 1D projected target dataset.
More recently an efficient framework for colour transfer was proposed for registering directly multidimensional Gaussian mixtures capturing target and source datasets using the $\mathcal{L}_2$ distance  \cite{GroganCVIU19}. Grogan \& Dahyot $\mathcal{L}_2$ framework  \cite{GroganCVIU19} relies on a parametric formulation with Thin Plate Splines (TPS) of the mapping function $\phi:\mathbb{R}^3\rightarrow \mathbb{R}^3$ that pushes  source image colours toward the target image ones.  
Our recent research efforts have extended IDT-SWD projection based algorithms using image patches  as samples in  higher dimensional spaces than the RGB triplets in $\mathbb{R}^3$ \cite{Alghamdi2019}, taking advantage of  correspondences between source and target patches \cite{alghamdi2020patch} and replacing OT 1D solution by the Nadaraya-Watson estimator \cite{alghamdi2020iterative}.
This paper proposes to extend this projective based strategy using the more robust  $\mathcal{L}_2$ distance \cite{scott2001parametric,JianPAMI2011} to infer mapping functions suitable with and without correspondences \cite{grogan2017user,GroganCVIU19}.  Using the 1D projective strategy also allows us to tackle the inherent limitation of TPS parametric formulation  that does not scale well in high dimensional spaces \cite{GroganCVIU19}.

\section{The Proposed Method}
\label{sec:method}

Our  Sliced $\mathcal{L}_2$ Distance (SL2D) approach for distribution transfer  is inspired by the projection-based approach utilized  by the  IDT \cite{Pitie_CVIU2007} and SWD \cite{BonneelJMIV2015} algorithms that have been originally  proposed to solve OT in $N$-dimensional spaces.

\begin{figure}[!h]
  \centering
  \includegraphics[width=.8\linewidth]{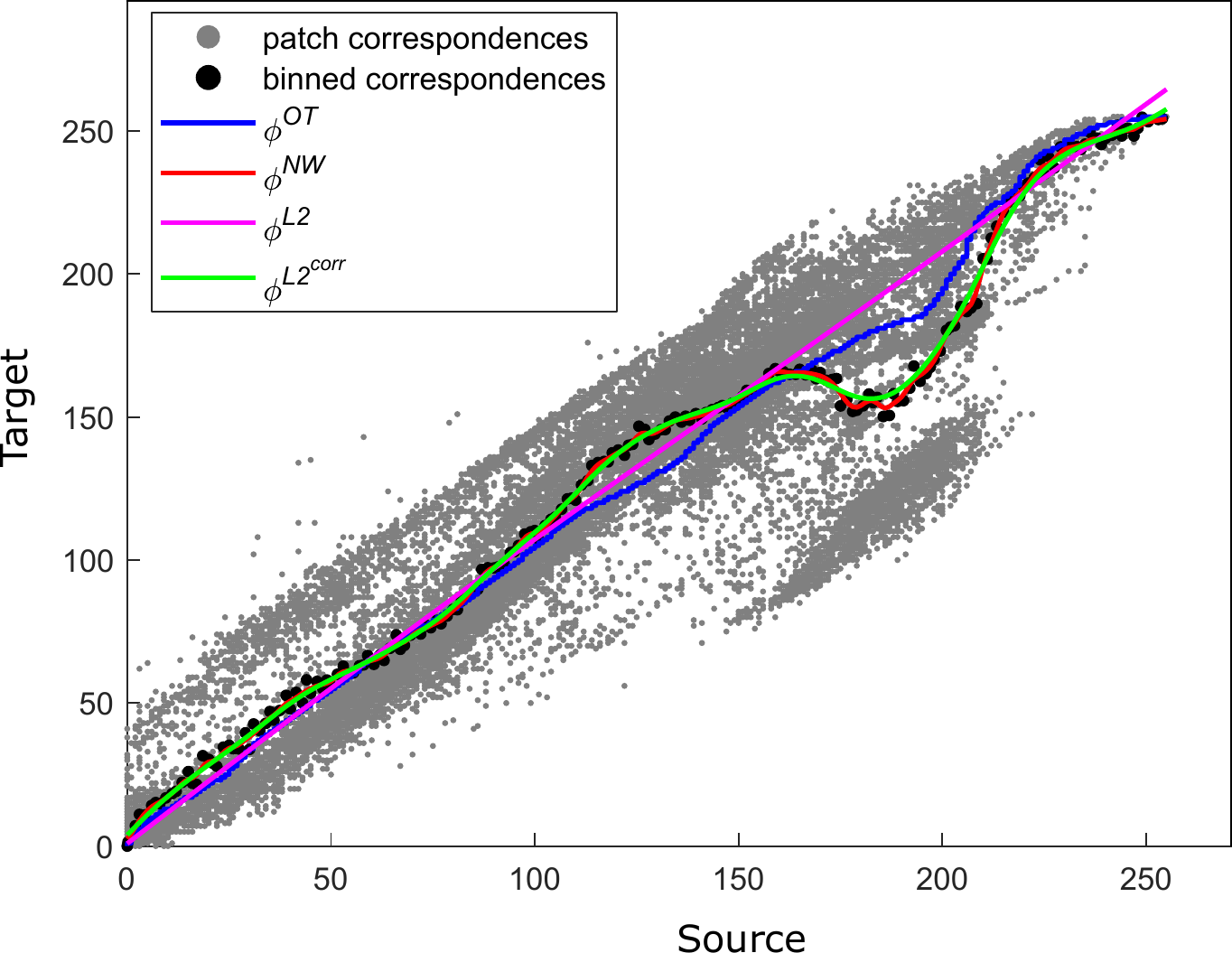}
  \caption{Mappings $\phi^{{\mathcal{L}_2}^{corr}}$, $\phi^{{\mathcal{L}_2}}$, in comparison to $\phi^{NW}$ \cite{alghamdi2020iterative} and  $\phi^{OT}$ \cite{Pitie_CVIU2007}.}
  \label{fig:L2vsothers}
\end{figure}

\subsection{$\mathcal{L}_2$ based solution vs Optimal Transport solution in 1D}
\label{sec:SL2D}

IDT algorithm \cite{Pitie_CVIU2007}  projects two multidimensional independent datasets $\{ x_i \}_{i=1}^n $ and $\{ y_j \}_{j=1}^m$ sampled for two random vectors $x\in \mathbb{R}^N$ and $y\in \mathbb{R}^N$ with respective distributions $f_x$ and $g_y$,    to 1D subspace. 
This projection creates two 1D datasets  $\{ u_i \}_{i=1}^n $ and $\{ v_j \}_{j=1}^m$ with corresponding marginals $f_u$ and $g_v$ whose cumulative distributions $F_u$ and $G_v$ are matched using the 1D optimal transport solution $\phi^{OT}(u)=G_v^{-1}\circ F_u(u)$ (cf. Fig. \ref{fig:L2vsothers}). We propose to replace the non-parametric $\phi^{OT}(u)$ by the following parametric non-rigid 1D transformation model:  
\begin{equation}
\phi^{\mathcal{L}_2}_{\theta}(u)=c_0+c_1\ u+\sum_{l=1}^{r} w_l \;\varphi ( {\lVert {u-u_l}\rVert}_2  )
\label{eq:ch7phi} 
\end{equation}
 $\{u_l\}_{l=1}^r$ are the control points, $\varphi ( {\lVert {u-u_l}\rVert}_2  ) = {\lVert {u-u_l}\rVert}_2 $ is the radial basis function, and $\theta =  \lbrace c_0,c_1,w_1,w_2,...,w_r \rbrace $ the parameters that control the transformation estimated with:
\begin{equation}
\hat{\theta}=\arg\min_{\theta} \; \bigg[ \;
\mathcal{L}_2(f_{u|\theta},g_v)=\|f_{u|\theta}-g_v\|^2 
\; \bigg]
\label{obtTheta}
\end{equation}
The probability density function (pdf) $f_{u|\theta}$ is a 1D Gaussian mixture model fitted to source samples $\{ \phi^{\mathcal{L}_2}_{\theta}(u_i) \}_{i=1}^n $, and the pdf  $g_v$  is a 1D Gaussian mixture fitted to the target samples $\{ v_j \}_{j=1}^m$.
 We experimented with different numbers $r$ of control points and we found  $r=125$ control points on regular intervals spanning the entire range of the 1D projected dataset give best results. As a consequence the dimension of the latent space that needs to be explored when estimating $\theta$ is:
$$\text{dim}(\theta)= (125 \times 1) + 1 + 1 =127  $$
with $\text{dim}(u_l)=1$, $\text{dim}(c_0)=1$, and $\text{dim}(c_1)=1$ (Eq.~\ref{eq:ch7phi}).

\subsection{Optimisation}

\begin{figure}[!h]
  \centering
  \includegraphics[width=.7\linewidth]{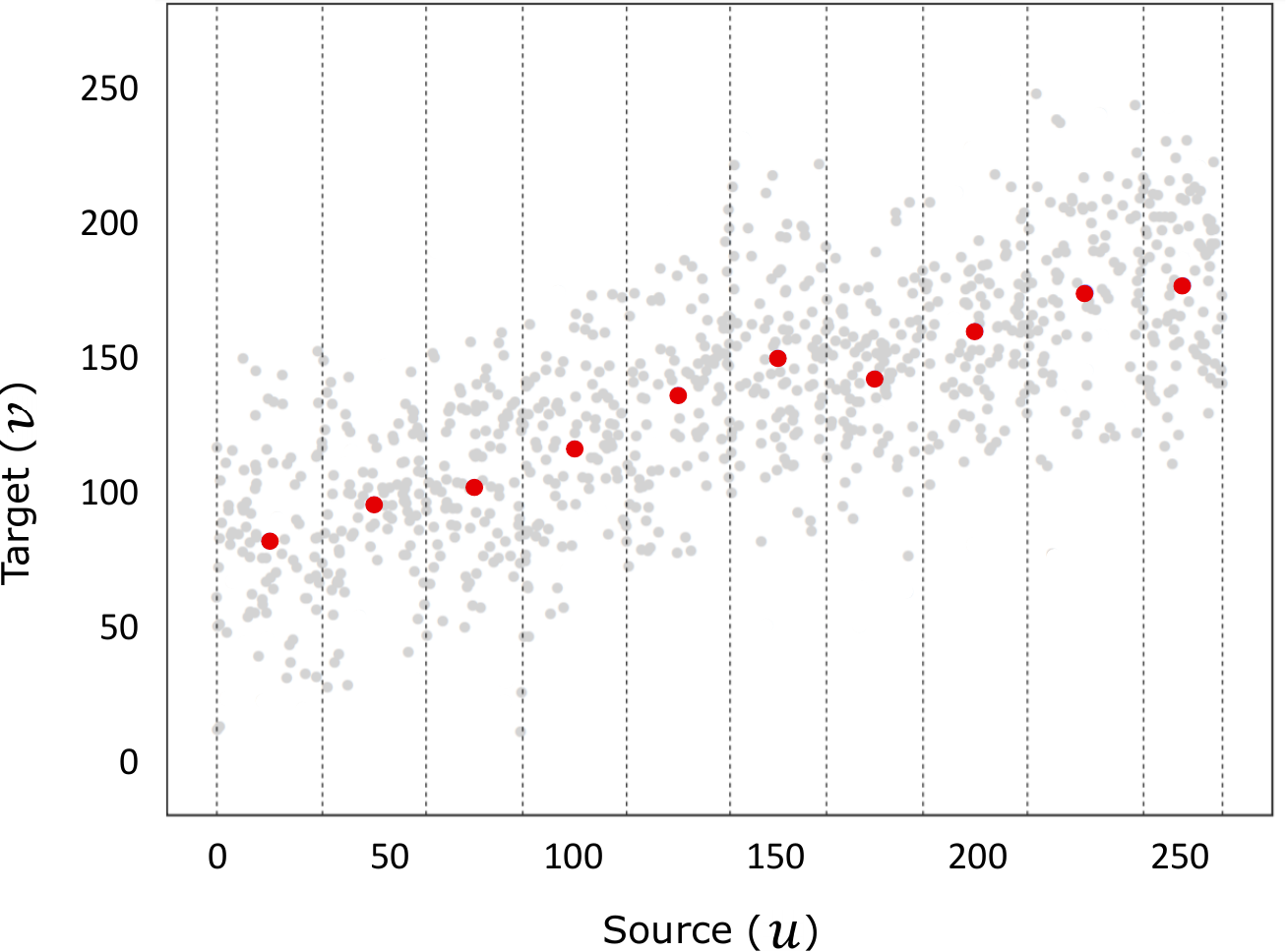}
  \caption{An illustration of creating binned correspondences $\{(\bar{u}_i,\bar{v}_i)\}_{i=1}^K$ from $\{(u_i,v_i)\}_{i=1}^q$. The red dots  representing the mean of the corresponding target values for the source observations falling in each bin. }
  \label{binned_scatterPlot}
\end{figure}

Following \cite{GroganCVIU19}, when no correspondences are available between source and target datasets, K-means algorithm is applied in $\mathbb{R}^N$ to reduce their cardinalities $n$ and $m$   to cardinality $K$. The projections of these K-means in 1D subspace define the  means of the  Gaussian Mixture models $f_u$ and $g_v$. 
A data-driven bandwidth  value $h$ is selected to control the isotropic variances of the Gaussians  \cite{JianPAMI2011}.
When $K$ correspondences in 1D  are available $\lbrace (u_k,v_k)\rbrace_{k=1,\cdots,K}$,  the binned correspondences procedure proposed in  \cite{alghamdi2020iterative} is used (c.f. Fig.~\ref{binned_scatterPlot}).

In order to estimate the parameters of the transformation $\phi_{\theta}$ in Eq.~(\ref{eq:ch7phi}) that minimizes the cost function in  Eq.~(\ref{obtTheta}), we used Quasi-Newton method \cite{QuasiNewton}, which is a gradient-based numerical optimization technique.

\subsection{Convergence }
\begin{figure}[!h]
  \centering
  \includegraphics[width=\linewidth]{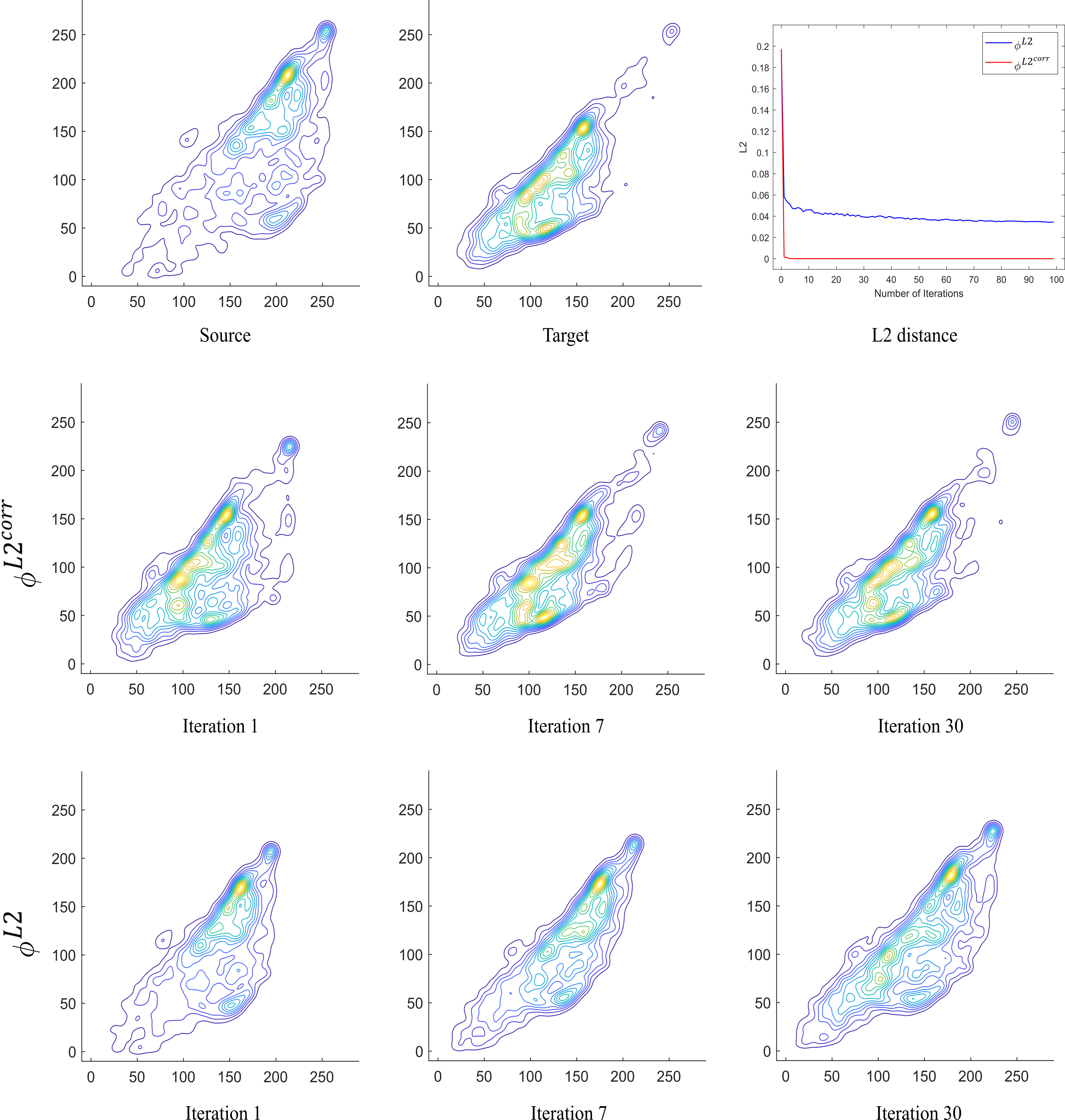}
  \caption{Example of pdf of the transferred source patches projected in 2D space (RG). The patch size chosen is $1\times 1$ and  only the colour information is used $N=3$ (space RGB). The standard $\mathcal{L}_2$ distance is computed after each iteration to illustrate the convergence of the source distribution to the target one by our transfer methods. We  note that $\phi^{\mathcal{L}_2}$ at iteration 30 is not yet matching the target distribution in comparison to  $\phi^{{\mathcal{L}_2}^{corr}}$ that is able to match the target and converge faster by iteration 7.}
  \label{fig:pdfTransfer}
\end{figure}

The  $\mathcal{L}_2$  distance between the $N$-dimensional source and target probability  distributions \cite{JianPAMI2011} is used as a measure to quantify how well the transformed distribution $f$ matches
the target pdf $g$ after each iteration $k$  of the algorithm. 
 Figure~\ref{fig:pdfTransfer} illustrates several iterations $k$ of our algorithm visualized in 2D space,  using correspondences ($\phi^{{\mathcal{L}_2}^{corr}}$) and without correspondences ($\phi^{\mathcal{L}_2}$). As we can see from the figure,   $\phi^{\mathcal{L}_2}$ at iteration 30 is not yet matching the target distribution in comparison to  $\phi^{{\mathcal{L}_2}^{corr}}$ that is able to match the target and converge faster by iteration 7.

\subsection{Interpolation between solutions }

\begin{figure*}
  \centering
  \includegraphics[width=\linewidth]{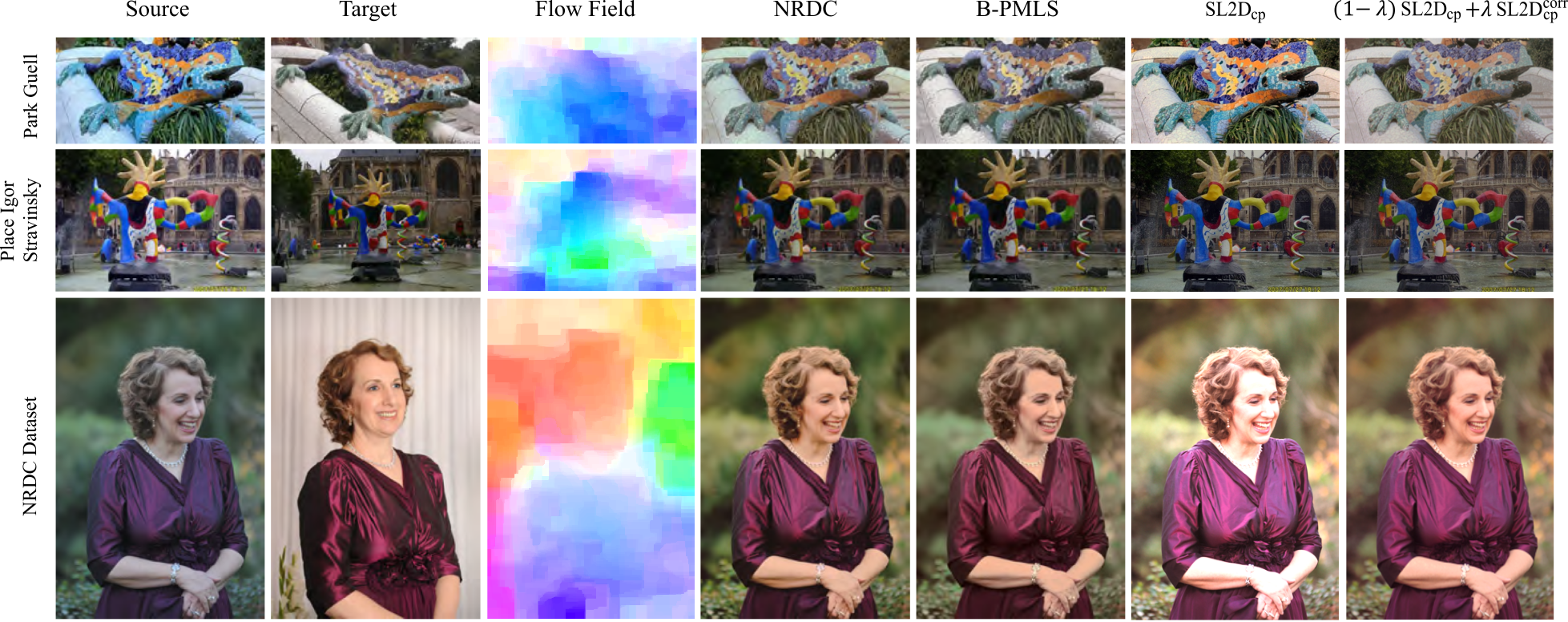}
  \caption{Colour transfer results of interpolating between our solutions \texttt{SL2D\textsupsub{corr}{cp}} and \texttt{SL2D\textsupsub{}{cp}}  with comparisons to the colour transfer results in \cite{HWANG20191,HaCohen2011}.}
  \label{fig:newData}
\end{figure*}

New transformations can be interpolated in each iteration between the solutions $\phi^{{\mathcal{L}_2}^{corr}}$ when taking into account the correspondences, and $\phi^{\mathcal{L}_2}$ without correspondences to tackle semi-supervised situations where correspondences are only partially available  \cite{grogan2017user,GroganCVIU19}:
\begin{equation}
\forall \; \lambda \in [0,1],\;\; \phi_{\lambda}^{\mathcal{L}_2^{int}}= (1-\lambda)\ \phi^{\mathcal{L}_2^{corr}}_{\hat{\theta}_1 } +\lambda \ \phi^{\mathcal{L}_2}_{\hat{\theta}_2} 
\end{equation}
This strategy  can be useful when no correspondence can be found locally between areas of the target and source images (e.g. non-overlapped areas or occlusions, Figure~\ref{fig:newData}).


\section{Experimental Assessment}
 \label{ch7:sec:experiments}
 
Quantitative  evaluations have been carried out with our techniques: the 1\ts{st}  without correspondences using colour patches only (\texttt{SL2D\textsupsub{}{c}}), the 2\ts{nd} without correspondences using colour patches with pixel  location  information 
(\texttt{SL2D\textsupsub{}{cp}}), the 3\ts{rd} with correspondences using colour patches only (\texttt{SL2D\textsupsub{corr}{c}}), and the 4\ts{th} with correspondences using colour patches with  pixel  location   (\texttt{SL2D\textsupsub{corr}{cp}}). 
We compare our methods with different state of the art colour transfer methods  noted by \texttt{B-PMLS} \cite{HWANG20191}, \texttt{L2} \cite{GroganCVIU19}, \texttt{GPS/LCP} and \texttt{FGPS/LCP}  \cite{bellavia2018dissecting}, \texttt{PMLS} \cite{hwang2014color}, \texttt{IDT} \cite{Pitie_CVIU2007}, 
\texttt{PCT\_OT} \cite{alghamdi2020patch}, \texttt{OT\_NW} \cite{alghamdi2020iterative} and \texttt{INWDT} \cite{alghamdi2020iterative}.

 Hwang et al. dataset \cite{hwang2014color} is used for evaluation and it includes registered pairs of images (source and target) taken with different cameras, different in-camera settings, and different illuminations and recolouring styles.
Results shown for \texttt{PMLS} \cite{hwang2014color} and \texttt{B-PMLS} \cite{HWANG20191} were provided by the authors. Two other recent techniques  \cite{Xia2017,park2016} that account for  correspondences were also tested for comparison  but are not reported as \texttt{PMLS} has been shown to outperforms these two \cite{GroganCVIU19}. We have also compared our approach with \cite{Bonneel2016,Ferradans2013} that do not  account for correspondences, however Grogan et al. \cite{GroganCVIU19} already reported that  \texttt{IDT}   is superior, hence \texttt{IDT} is the one reported here for ease of comparison.

We use  the RGB colour space and  we found a patch size of $3 \times 3$ captures enough of a pixel's neighbourhood.  For our \texttt{SL2D\textsupsub{}{cp}} and \texttt{SL2D\textsupsub{corr}{cp}} versions, each pixel is represented by its 3D RGB colour values  and its 2D pixel position (i.e 5D). The patches with combined colour and spatial features create a vector in 45 dimensions ($N=5 \times 3\times 3=45$). For \texttt{SL2D\textsupsub{}{c}} and \texttt{SL2D\textsupsub{corr}{c}}, pixel position is not accounted for, and only RGB colours are used, which create patch vectors in 27 dimensions  ($N=3 \times 3\times 3=27$).

\begin{figure}[!h]
\centering
\begin{tabular}{c}
\includegraphics[width=.9\linewidth]{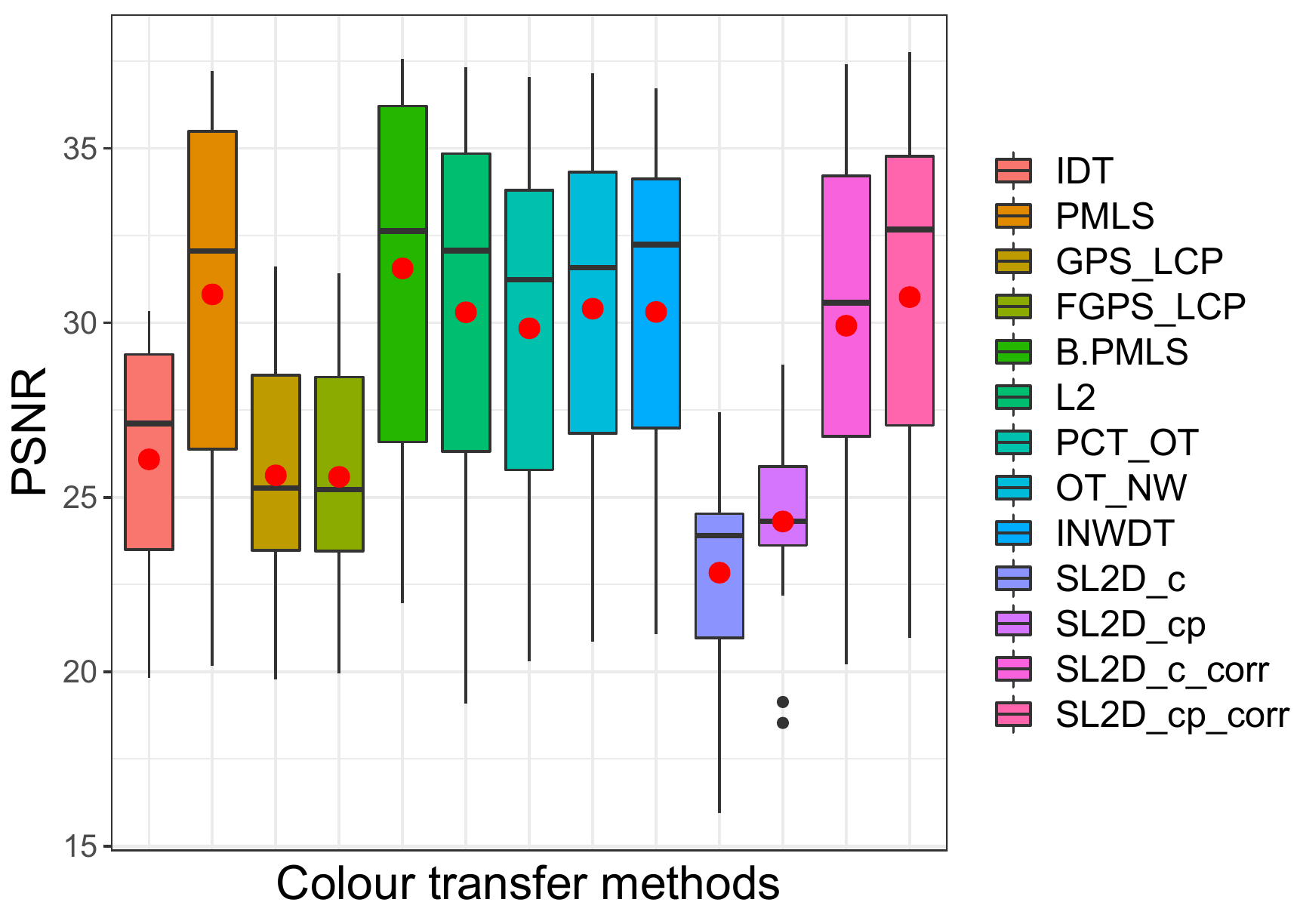}
\end{tabular}
\caption{Comparing our algorithms  \texttt{SL2D\textsupsub{}{c}}, \texttt{SL2D\textsupsub{}{cp}}, \texttt{SL2D\textsupsub{corr}{c}} and \texttt{SL2D\textsupsub{corr}{cp}} with the state of the art colour transfer methods using PSNR metric  \cite{salomon2004data} (higher values are better, best viewed in colour and zoomed in).}
\label{fig:ch7_psnr_plot}
\end{figure}

\begin{figure}[!h]
\centering
\begin{tabular}{c}
\includegraphics[width = .9\linewidth]{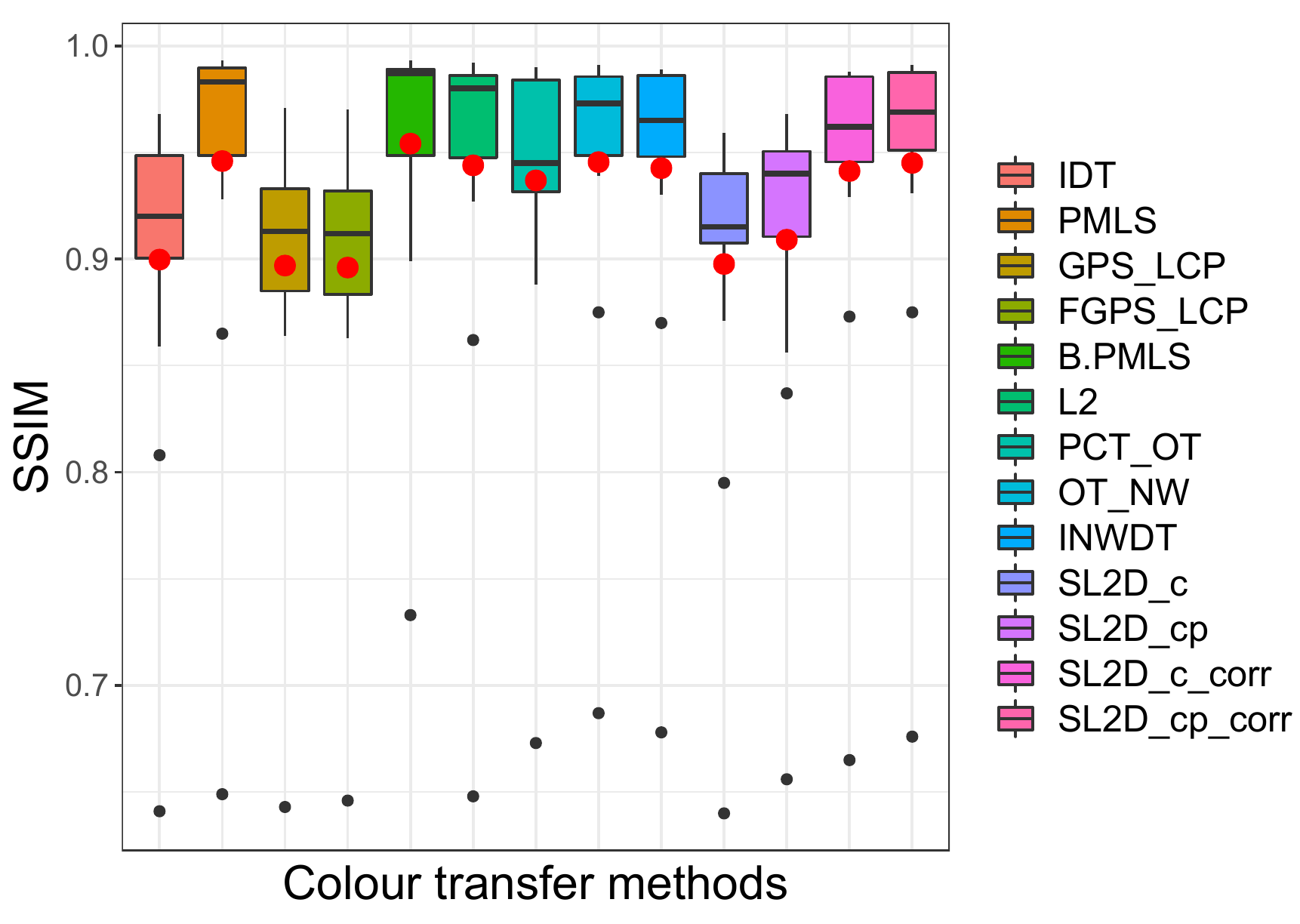}
\end{tabular}
\caption{Comparing \texttt{SL2D\textsupsub{}{c}}, \texttt{SL2D\textsupsub{}{cp}}, \texttt{SL2D\textsupsub{corr}{c}} and \texttt{SL2D\textsupsub{corr}{cp}} with the state of the art colour transfer methods using SSIM metric \cite{wang2004image} (higher values are better, best viewed in colour and zoomed in).}
\label{fig:ch7_ssim_plot}
\end{figure}

\begin{figure}[!h]
\centering
\begin{tabular}{c}
\includegraphics[width = .9\linewidth]{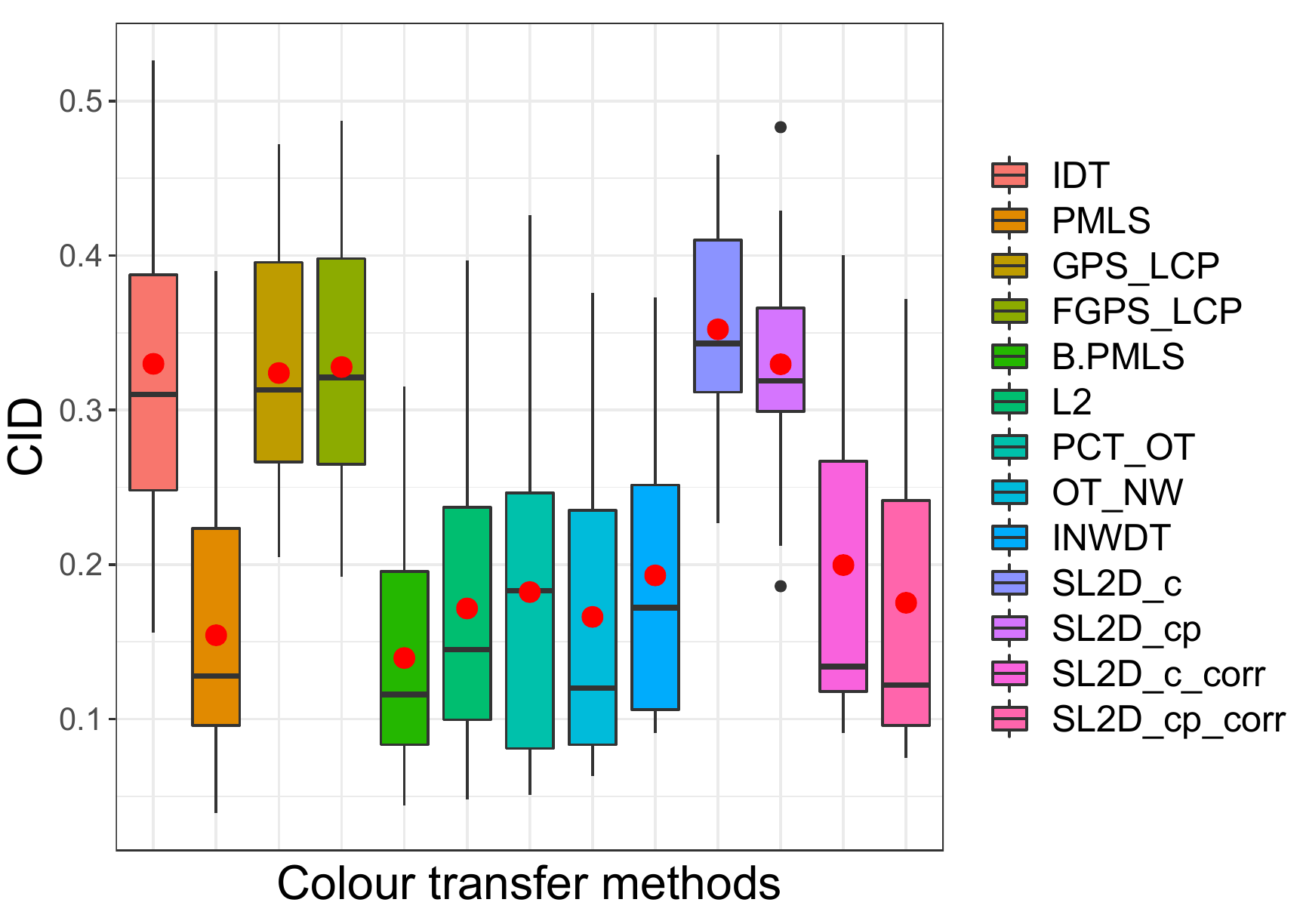}
\end{tabular}
\caption{Comparing \texttt{SL2D\textsupsub{}{c}}, \texttt{SL2D\textsupsub{}{cp}}, \texttt{SL2D\textsupsub{corr}{c}} and \texttt{SL2D\textsupsub{corr}{cp}} with the state of the art colour transfer methods using CID metric \cite{preiss2014color} (lower values are better, best viewed in colour and zoomed in).}
\label{fig:ch7_cid_plot}
  \end{figure}

\begin{figure}[!h]
\centering
\begin{tabular}{c}
\includegraphics[width = .9\linewidth]{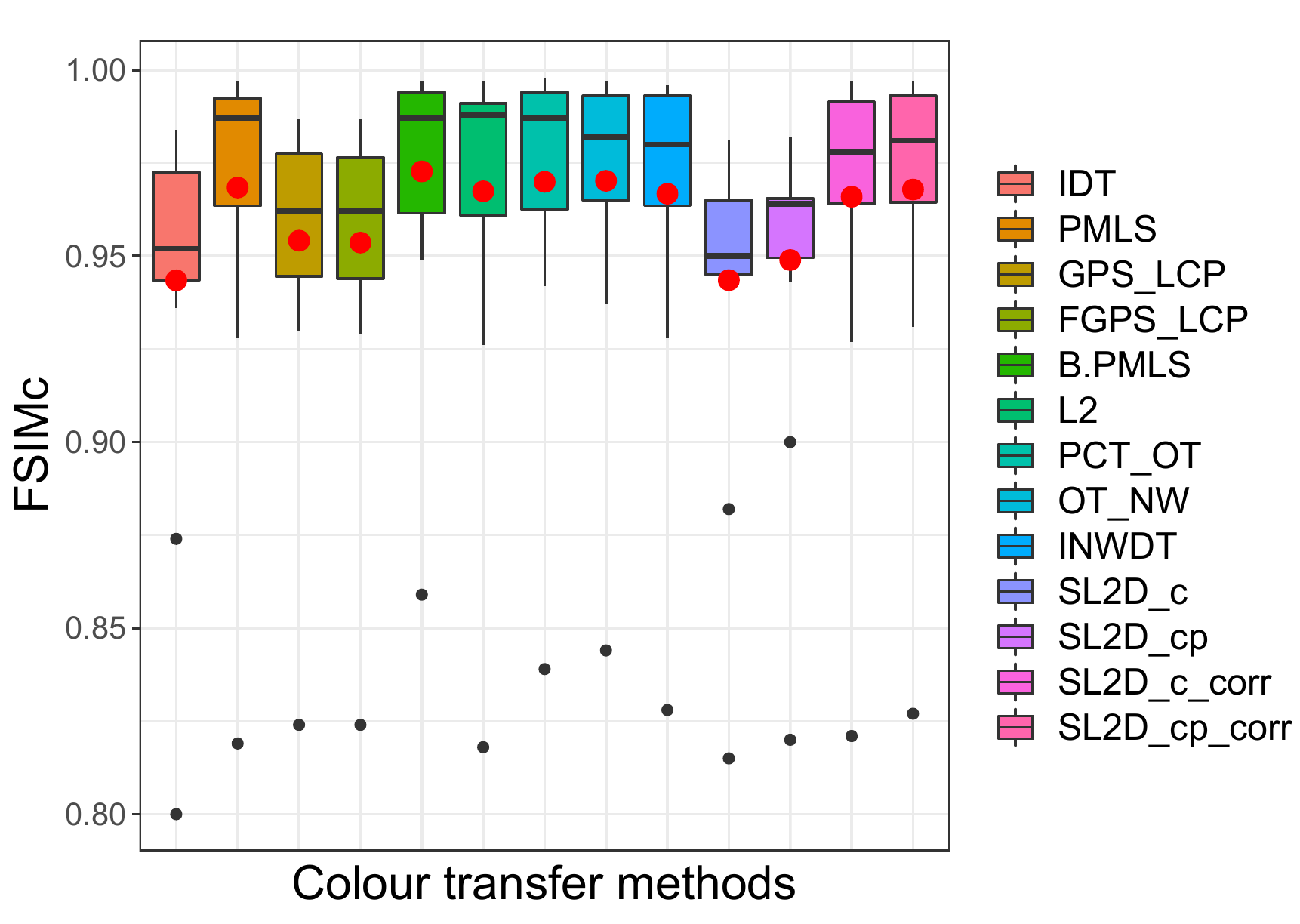}
\end{tabular}
\caption{Comparing \texttt{SL2D\textsupsub{}{c}}, \texttt{SL2D\textsupsub{}{cp}}, \texttt{SL2D\textsupsub{corr}{c}} and \texttt{SL2D\textsupsub{corr}{cp}} with the state of the art colour transfer methods using FSIMc metric \cite{zhang2011fsim} (higher values are better, best viewed in colour and zoomed in).}
\label{fig:ch7_fsimc_plot}
\end{figure}

\begin{figure*}
  \centering
  \includegraphics[width=.93\linewidth,keepaspectratio]{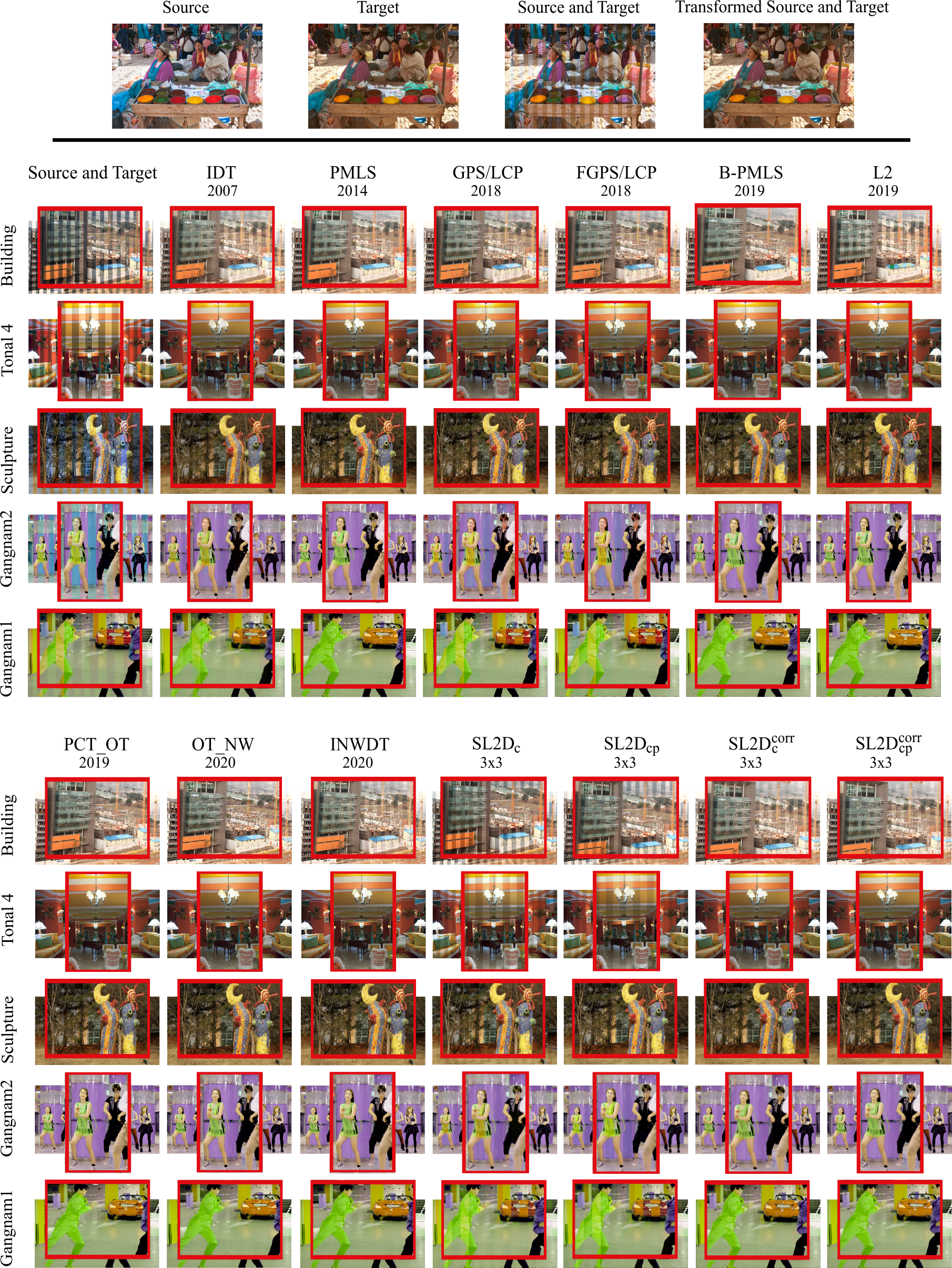}
  \caption{A close up look at some of the results generated using the \texttt{IDT} \cite{Pitie_CVIU2007}, \texttt{PMLS} \cite{hwang2014color}, \texttt{GPS/LCP} and \texttt{FGPS/LCP}  \cite{bellavia2018dissecting},  \texttt{B-PMLS} \cite{HWANG20191}, \texttt{L2} \cite{GroganCVIU19}, \texttt{PCT\_OT} \cite{Alghamdi2019}, \texttt{OT\_NW} \cite{alghamdi2020patch} , \texttt{INWDT} \cite{alghamdi2020iterative} and our algorithms using correspondences (\texttt{SL2D\textsupsub{corr}{c}} and \texttt{SL2D\textsupsub{corr}{cp}}) and without using correspondences (\texttt{SL2D\textsupsub{}{c}} and \texttt{SL2D\textsupsub{}{cp}}) - best viewed in colour and zoomed in.}
  \label{fig:L2_qualitative_strips}
\end{figure*}

The numerical results for PSNR, SSIM, CID and FSIMc   are shown in  box plots shown in Figures~\ref{fig:ch7_psnr_plot} to \ref{fig:ch7_fsimc_plot} (the means shown as red dots in the plots, and the medians shown as horizontal black lines).  Sliced $\mathcal{L}_2$ with correspondences  (\texttt{SL2D\textsupsub{corr}{c}} and \texttt{SL2D\textsupsub{corr}{cp}}) significantly outperforms the Sliced $\mathcal{L}_2$ solutions without correspondences (\texttt{SL2D\textsupsub{}{c}} and \texttt{SL2D\textsupsub{}{cp}}). 
Incorporating colour with spatial information (\texttt{SL2D\textsupsub{corr}{cp}}) improves the performance over using colour information only (\texttt{SL2D\textsupsub{corr}{c}}). Moreover, the  iterative projection approach with $\mathcal{L}_2$   (colour only \texttt{SL2D\textsupsub{corr}{c}} and combined colour and position \texttt{SL2D\textsupsub{corr}{cp}}) outperform the iterative projection approach with  OT  solution (IDT algorithm). 
The medians reported  for techniques  
 \texttt{B-PMLS}, \texttt{L2}, \texttt{PMLS}, \texttt{OT\_NW}, \texttt{INWDT}, \texttt{SL2D\textsupsub{corr}{c}} and  \texttt{SL2D\textsupsub{corr}{cp}} are not statistically different ($ 95\%$ confidence level) indicating equivalent quantitative performance.
 For qualitative performance, see Fig. \ref{fig:L2_qualitative_strips} showing SL2D performance against State of the Art. 
Multiple projections and 1D pdf registrations in our algorithm can be performed in parallel in the same fashion as in IDT and SWD.
 
\section{Conclusion}
We have introduced the Sliced $\mathcal{L}_2$ distance for registering two high dimensional probability density functions,  that allows  correspondences  to be taken into account when these are available. 
Our SL2D technique applied to colour transfer  extends Grogan et al. $\mathcal{L}_2$ approach   \cite{grogan2017user,GroganCVIU19} to higher dimensional spaces, and performs well against the state of the art. Future work will look at applying 
SL2D for shape registration \cite{JianPAMI2011,GROGAN2018452}.

\bibliographystyle{IEEEtran}
\bibliography{thesis}

\begin{thebibliography}{10}
\providecommand{\url}[1]{#1}
\csname url@samestyle\endcsname
\providecommand{\newblock}{\relax}
\providecommand{\bibinfo}[2]{#2}
\providecommand{\BIBentrySTDinterwordspacing}{\spaceskip=0pt\relax}
\providecommand{\BIBentryALTinterwordstretchfactor}{4}
\providecommand{\BIBentryALTinterwordspacing}{\spaceskip=\fontdimen2\font plus
\BIBentryALTinterwordstretchfactor\fontdimen3\font minus
  \fontdimen4\font\relax}
\providecommand{\BIBforeignlanguage}[2]{{%
\expandafter\ifx\csname l@#1\endcsname\relax
\typeout{** WARNING: IEEEtran.bst: No hyphenation pattern has been}%
\typeout{** loaded for the language `#1'. Using the pattern for}%
\typeout{** the default language instead.}%
\else
\language=\csname l@#1\endcsname
\fi
#2}}
\providecommand{\BIBdecl}{\relax}
\BIBdecl

\bibitem{peyre2019computational}
G.~Peyr{\'e} and M.~Cuturi, ``Computational optimal transport: With
  applications to data science,'' \emph{Foundations and Trends{\textregistered}
  in Machine Learning}, vol.~11, no. 5-6, pp. 355--607, 2019.

\bibitem{TanakaNIPS2019}
A.~Tanaka, ``Discriminator optimal transport,'' in \emph{Adv. Neural Inf.
  Process. Syst.}, 2019, pp. 6816--6826.

\bibitem{NIPS2019_projPursuit}
C.~Meng, Y.~Ke, J.~Zhang, M.~Zhang, W.~Zhong, and P.~Ma, ``Large-scale optimal
  transport map estimation using projection pursuit,'' in \emph{Adv. Neural
  Inf. Process. Syst.}, 2019, pp. 8118--8129.

\bibitem{NIPS2019_Subspace_Detours}
B.~Muzellec and M.~Cuturi, ``Subspace detours: Building transport plans that
  are optimal on subspace projections,'' in \emph{Adv. Neural Inf. Process.
  Syst.}, 2019, pp. 6917--6928.

\bibitem{Pitie_CVIU2007}
F.~Pitie, A.~C. Kokaram, and R.~Dahyot, ``Automated colour grading using colour
  distribution transfer,'' \emph{Comput. Vis. Image Underst.}, vol. 107, no.~1,
  pp. 123 -- 137, 2007.

\bibitem{Pitie_ICCV2005}
F.~{Pitie}, A.~C. {Kokaram}, and R.~{Dahyot}, ``N-dimensional probability
  density function transfer and its application to color transfer,'' in
  \emph{IEEE Int. Conf. Comput. Vis. (ICCV)}, vol.~2, 2005, pp. 1434--1439.

\bibitem{BonneelJMIV2015}
N.~Bonneel, J.~Rabin, G.~Peyr{\'e}, and H.~Pfister, ``Sliced and radon
  wasserstein barycenters of measures,'' \emph{J. Math. Imaging Vis.}, vol.~51,
  no.~1, pp. 22--45, Jan 2015.

\bibitem{GroganCVIU19}
M.~Grogan and R.~Dahyot, ``L2 divergence for robust colour transfer,''
  \emph{Comput. Vis. Image Underst.}, vol. 181, pp. 39 -- 49, 2019.

\bibitem{Alghamdi2019}
H.~{Alghamdi}, M.~{Grogan}, and R.~{Dahyot}, ``Patch-based colour transfer with
  optimal transport,'' in \emph{Proc. Eur. Signal Process. Conf. (EUSIPCO)},
  Sep. 2019, pp. 1--5.

\bibitem{alghamdi2020patch}
H.~Alghamdi and R.~Dahyot, ``Patch based colour transfer using sift flow,'' in
  \emph{Proc. Irish Mach. Vis. Image Process. conf. (IMVIP)}, 2020.

\bibitem{alghamdi2020iterative}
------, ``Iterative nadaraya-watson distribution transfer for colour grading,''
  in \emph{Proc. IEEE Int. Workshop Multimed. Signal Process. (MMSP)}.\hskip
  1em plus 0.5em minus 0.4em\relax Ieee, 2020, pp. 1--6.

\bibitem{scott2001parametric}
D.~W. Scott, ``Parametric statistical modeling by minimum integrated square
  error,'' \emph{Technometrics}, vol.~43, no.~3, pp. 274--285, 2001.

\bibitem{JianPAMI2011}
B.~{Jian} and B.~C. {Vemuri}, ``Robust point set registration using gaussian
  mixture models,'' \emph{IEEE Trans. Pattern Anal. Mach. Intell.}, vol.~33,
  no.~8, pp. 1633--1645, 2011.

\bibitem{grogan2017user}
M.~Grogan, R.~Dahyot, and A.~Smolic, ``User interaction for image recolouring
  using l2,'' in \emph{Proc. Eur. Conf. on Visual Media Production (CVMP)},
  2017, p.~10.

\bibitem{QuasiNewton}
D.~F. Shanno, ``Conditioning of quasi-newton methods for function
  minimization,'' \emph{Math. Comp.}, vol.~24, no. 111, pp. 647--656, 1970.

\bibitem{HWANG20191}
Y.~Hwang, J.-Y. Lee, I.~S. Kweon, and S.~J. Kim, ``Probabilistic moving least
  squares with spatial constraints for nonlinear color transfer between
  images,'' \emph{Comput. Vis. Image Underst.}, vol. 180, pp. 1 -- 12, 2019.

\bibitem{bellavia2018dissecting}
F.~{Bellavia} and C.~{Colombo}, ``Dissecting and reassembling color correction
  algorithms for image stitching,'' \emph{IEEE Trans. Image Process.}, vol.~27,
  no.~2, pp. 735--748, 2018.

\bibitem{hwang2014color}
Y.~{Hwang}, J.~{Lee}, I.~S. {Kweon}, and S.~J. {Kim}, ``Color transfer using
  probabilistic moving least squares,'' in \emph{Proc. IEEE Comput. Vis.
  Pattern Recognit. (CVPR)}, 2014, pp. 3342--3349.

\bibitem{Xia2017}
M.~Xia, J.~Y. Renping, X.~M. Zhang, and J.~Xiao, ``Color consistency correction
  based on remapping optimization for image stitching,'' in \emph{IEEE Int.
  Conf. Comput. Vis. Workshops}, Oct 2017, pp. 2977--2984.

\bibitem{park2016}
J.~Park, Y.~Tai, S.~N. Sinha, and I.~S. Kweon, ``Efficient and robust color
  consistency for community photo collections,'' in \emph{Proc. IEEE Comput.
  Vis. Pattern Recognit. (CVPR)}, June 2016, pp. 430--438.

\bibitem{Bonneel2016}
N.~Bonneel, G.~Peyr\'{e}, and M.~Cuturi, ``Wasserstein barycentric coordinates:
  Histogram regression using optimal transport,'' \emph{ACM Trans. Graph.
  (TOG)}, vol.~35, no.~4, Jul. 2016.

\bibitem{Ferradans2013}
S.~Ferradans, N.~Papadakis, J.~Rabin, G.~Peyr{\'e}, and J.-F. Aujol,
  ``Regularized discrete optimal transport,'' in \emph{Int. Conf. on Scale
  Space and Variational Methods in Computer Vision}.\hskip 1em plus 0.5em minus
  0.4em\relax Springer Berlin Heidelberg, 2013, pp. 428--439.

\bibitem{salomon2004data}
D.~Salomon, \emph{Data Compression: The Complete Reference}.\hskip 1em plus
  0.5em minus 0.4em\relax Springer New York, 2004.

\bibitem{wang2004image}
{Zhou Wang}, A.~C. {Bovik}, H.~R. {Sheikh}, and E.~P. {Simoncelli}, ``Image
  quality assessment: from error visibility to structural similarity,''
  \emph{IEEE Trans. Image Process.}, vol.~13, no.~4, pp. 600--612, 2004.

\bibitem{preiss2014color}
J.~{Preiss}, F.~{Fernandes}, and P.~{Urban}, ``Color-image quality assessment:
  From prediction to optimization,'' \emph{IEEE Trans. Image Process.},
  vol.~23, no.~3, pp. 1366--1378, 2014.

\bibitem{zhang2011fsim}
L.~{Zhang}, L.~{Zhang}, X.~{Mou}, and D.~{Zhang}, ``Fsim: A feature similarity
  index for image quality assessment,'' \emph{IEEE Trans. Image Process.},
  vol.~20, no.~8, pp. 2378--2386, 2011.

\bibitem{HaCohen2011}
Y.~HaCohen, E.~Shechtman, D.~B. Goldman, and D.~Lischinski, ``Non-rigid dense
  correspondence with applications for image enhancement,'' \emph{ACM Trans.
  Graph. (TOG)}, vol.~30, no.~4, Jul. 2011.

\bibitem{GROGAN2018452}
M.~Grogan and R.~Dahyot, ``Shape registration with directional data,''
  \emph{Pattern Recognition}, vol.~79, pp. 452 -- 466, 2018.

\end{thebibliography}

\end{document}